\documentclass[letterpaper]{article} % DO NOT CHANGE THIS
\usepackage{aaai2026}  % DO NOT CHANGE THIS
\usepackage{times}  % DO NOT CHANGE THIS
\usepackage{helvet}  % DO NOT CHANGE THIS
\usepackage{courier}  % DO NOT CHANGE THIS
\usepackage[hyphens]{url}  % DO NOT CHANGE THIS
\usepackage{graphicx} % DO NOT CHANGE THIS
\usepackage{natbib}  % DO NOT CHANGE THIS AND DO NOT ADD ANY OPTIONS TO IT
\usepackage{caption} % DO NOT CHANGE THIS AND DO NOT ADD ANY OPTIONS TO IT
\usepackage{algorithm}
\usepackage{amsmath} % For mathematical equations
\usepackage{amssymb} % For mathematical symbols
\usepackage{algpseudocode} % For pseudocode commands
\usepackage{graphicx}
\usepackage{tabularx}
\usepackage{breqn}

\usepackage{booktabs}  
\usepackage{subcaption}
\newtheorem{prop}{Proposition}

\usepackage{newfloat}
\usepackage{listings}
\DeclareCaptionStyle{ruled}{labelfont=normalfont,labelsep=colon,strut=off} % DO NOT CHANGE THIS
\floatstyle{ruled}
\newfloat{listing}{tb}{lst}{}
\floatname{listing}{Listing}
\title{ML-Guided Primal Heuristics \\for Mixed Binary Quadratic Programs}
\author{
    Weimin Huang\textsuperscript{\rm 1},
    Natalie M. Isenberg\textsuperscript{\rm 2},
    Ján Drgoňa\textsuperscript{\rm 3},
    Draguna L Vrabie\textsuperscript{\rm 2},
    Bistra Dilkina\textsuperscript{\rm 1}
}
\affiliations{
     \textsuperscript{\rm 1}University of Southern California, USA\\
    \textsuperscript{\rm 2}Pacific Northwest National Laboratory, USA\\
    \textsuperscript{\rm 3}Johns Hopkins University, USA\\
    weiminhu@usc.edu, natalie.isenberg@pnnl.gov, jdrgona1@jh.edu, draguna.vrabie@pnnl.gov, dilkina@usc.edu
}

\usepackage{bibentry}
\begin{document}

\maketitle

\begin{abstract}
Mixed Binary Quadratic Programs (MBQPs) are an important and complex set of problems in combinatorial optimization. As solving large-scale combinatorial optimization problems is challenging, primal heuristics have been developed to quickly identify high-quality solutions within a short amount of time. Recently, a growing body of research has also used machine learning to accelerate solution methods for challenging combinatorial optimization problems. Despite the increasing popularity of these ML-guided methods, a large body of work has focused on Mixed-Integer Linear Programs (MILPs). 
MBQPs are challenging to solve due to the combinatorial complexity coupled with nonlinearities. This work proposes ML-guided primal heuristics for Mixed Binary Quadratic Programs (MBQPs) by adapting and extending existing work on ML-guided MILP solution prediction to MBQPs. We introduce a new neural network architecture for MBQP solution prediction and a new training data collection procedure. Moreover, we extend existing loss functions in solution prediction and propose to combine contrastive and weighted cross-entropy losses. We evaluate the methods on standard and real-world MBQP benchmarks and show that the developed ML-guided methods significantly outperform existing primal heuristics and state-of-the-art solvers. Furthermore, models trained with our proposed extension with combined losses outperform other ML-based methods adapted from MILPs and improve generalization in cross-regional inference on a real-world wind farm layout optimization problem.
\end{abstract}

\begin{links}
    %\link{Code}{https://github.com/pnnl-private/MBQP_heuristics}
    %\link{Extended version}{https://arxiv.org/abs/2604.23053}
\end{links}

\section{Introduction}
Mixed Binary Quadratic Programs (MBQPs) are discrete optimization problems with quadratic terms in the objective function subject to a set of linear constraints. MBQPs encode many important problems in Combinatorial Optimization (CO)  ~\citep{loiola2007survey,rebennack2024stable,kochenberger2005unconstrained} and cover a wide range of applications, including finance~\citep{parpas2006global}, machine learning~\citep{bertsimas2009algorithm}, as well as chemical~\citep{misener2013glomiqo} and energy systems~\citep{turner2014new}. 
A significant body of research on CO algorithms has focused on 
\emph{primal heuristics}, which are algorithms designed to find good feasible solutions quickly and without optimality guarantees \citep{berthold2014rens}. 

%However, there has been limited existing work on primal heuristics specifically designed for solving MBQPs - many primal heuristics were derived from algorithms for Mixed Integer Linear Programming (MILP) problems \citep{berthold2014rens,bonami2009feasibility}.

Despite development in solvers and heuristics, solving large-scale COs remains challenging. In recent years, Machine Learning (ML) has been proposed to accelerate solution methods for CO problems. Motivated by the fact that CO problems that share similar structures are solved repeatedly in many applications \citep{huang2024distributional,scavuzzo2024machine}, a growing body of research uses ML to guide algorithmic policies or to build new policies customized to instances that appear in specific applications. For example, \citet{han2023a,nair2020solving,pmlr-v235-huang24f,ding2020accelerating} propose ML-guided primal heuristics for Mixed Integer Linear Programs (MILPs), wherein they predict the optimal assignment for a subset of the variables. \citet{lee2024rl} have shown that ML-guided methods reach high-quality solutions faster than non-ML primal heuristics on MILPs. While prior work on ML-guided CO methods has shown success across multiple algorithmic components on many challenging CO problems, existing work in this area has mainly focused on MILPs. A small body of research has used ML to advance solution methods for general nonlinear programming problems \citep{bonami2018learning,bagga2023solving,ghaddar2023learning,ferber2023surco,tang2025learningoptimizemixedintegernonlinear}, but ML-guided methods in this space are not as well developed as in MILPs.
%OF WORK HAS BEEN USED TO ADVANCE MINLP SOLVING, BUT NOT AS WELL-DEVELOPED AS MILP. 
%SOLVER CONIFG. SURROGATE SOLVER. CONTINUOUS QUADRATIC PROGRAMS (CHRISTOPHER MORRIS, PASCAL)] 
% With this work, we aim to transfer this success to motion planning of autonomous systems.
%While prior work on ML-guided MILP solving has shown success across multiple algorithmic families (local search vs. tree search) and algorithmic components (tree search branching vs tree search node selection), here 

MBQPs are even more challenging to solve than MILPs due to the combinatorial nature \citep{magnanti1981combinatorial} coupled with nonlinearities. In this work, we develop ML-guided primal heuristics for MBQPs by adapting and extending existing work on ML-guided MILP solution methods. We adapt the Weighted Cross-Entropy-based and Contrastive Learning-based methods which are used in MILP solution prediction to MBQPs. To adapt to MBQPs, we propose a novel neural network architecture that extracts input features and a new data collection procedure that generates high-quality solutions as ground truth training data for large-scale MBQPs. Furthermore, we extend existing loss functions used in CO solution prediction and propose to combine Cross-Entropy and Contrastive losses. Computational results show that the adapted and extended ML methods outperform existing primal heuristics and state-of-the-art solvers on standard and real-world MBQP benchmarks. Furthermore, we show that solution prediction models trained with our proposed extended loss function outperform other ML-guided methods adapted from MILPs in in-domain testing and improve generalization in cross-regional inference on a real-world wind farm layout optimization problem.

\section{Background and Related Works}
\subsection{Mixed Binary Quadratic Programs}\label{background}
A Mixed Binary Quadratic Program (MBQP) with $n$ decision variables is defined as
\begin{equation}
\min x^T H x + c^T x \quad \text{s.t.} \;A x \leq b \; \text{and}\; x_j \in \{0,1\}, \forall j \in B \label{eq:mbqp}
\end{equation}
where $H \in  \mathbb{R}^{n \times n}$, $c \in \mathbb{R}^n$, $A \in  \mathbb{R}^{m\times n}$, and $b \in  \mathbb{R}^m$. $H$ is a real symmetric matrix that encodes quadratic terms in the objective function and is not necessarily positive semidefinite, allowing for nonconvex objective functions. $B \subseteq \{1,...,n\}$ is the set of binary decision variables. 

\paragraph{Solution Methods}
MBQPs are NP-hard in general \citep{pia2017mixed}. The Branch-and-Bound (BnB) algorithm is an exact tree search algorithm to solve MILPs, MBQPs and more general Mixed-Integer Nonlinear Programming (MINLP) problems. As large-scale MBQPs are challenging to solve with exact methods, a significant body of research has focused on 
\emph{primal heuristics}, which are algorithms designed to quickly identify high-quality feasible solutions for a given optimization problem without optimality guarantees \citep{berthold2014rens}. These heuristics typically involve solving a relaxation of the original problem and then creating a subproblem by fixing a subset of integer variables by rounding the relaxation values to the nearest integer values, such as RENS \citep{berthold2014rens}, Undercover \citep{berthold2014undercover}, and Relax-Search \citep{Huangsocs_2025}.

\subsection{Solution Prediction for MILPs}\label{nd}

Previous work on using ML to accelerate solving CO problems has been focused on Mixed Integer Linear Programming (MILP). An MILP can be viewed as the subclass of MBQPs in Eqn. (\ref{eq:mbqp}) where the quadratic term matrix $H$ is the zero matrix. The goal of an MILP is to find \( x\) such that \( c^T x \) is minimized, subject to \( Ax \leq b \) and integrality constraints $x_j \in \{0,1\}, \forall j \in B$. A large body of ML-guided primal heuristics for MILPs is based on predicting partial solutions  \citep{nair2020solving,han2023a,pmlr-v235-huang24f,ding2020accelerating}. 

\paragraph{Solution Prediction}
\citet{nair2020solving} and \citet{han2023a} use Weighted Cross-Entropy (WCE) loss to learn the probability distribution of the solution space of an MILP instance $M$. The goal is to learn from a set of multiple solutions, weighted by the quality of the solution. Specifically, for a solution $x$, the energy function $E(x;M)$ is defined as $c^Tx$ if $x$ is feasible, or $\infty$ otherwise, assuming minimization. Given $M$, the conditional distribution of a solution $x$ is modeled as 
\begin{equation}
		\begin{aligned}
			P(x|M) \equiv \frac{\text{exp}(-E(x;M))}{\sum_{x^\prime }\text{exp}(-E(x^\prime;M))}\\
		\end{aligned}\label{eq:predict:energy}
\end{equation}
, so that solutions with better objective values have a higher probability. The learning task is to train a model $p(x|\theta;M)$ parameterized by $\theta$ that approximates $p(x|M)$. To collect training data, \citet{nair2020solving} and \citet{han2023a} obtain the set of solutions by running state-of-the-art MILP solvers for a large amount of time. Instead of using WCE loss, \citet{pmlr-v235-huang24f} learn $p(x|\theta;M)$ using Contrastive Learning (CL). The CL-based method makes discriminative predictions by contrasting the positive samples (i.e., good solutions) and negative samples (i.e., bad solutions). Positive samples are obtained by running MILP solvers, similar to \citep{nair2020solving} and \citep{han2023a}. Negative samples are obtained by solving another MILP that searches for bad variable assignments within some Hamming distance of the good solutions.
%They train a Graph Convolution Network (GCN) to learn a $p_\theta(x|M)$ with learnable parameters $\theta$. The idea is to learn from a set of multiple assignments, not just the best ones. The learning task is to approximate $p(x|M)$ using a graph convolutional network. $P_\theta\left(x^{i,j};M_i\right)$ denotes the prediction from the GNN denoted as $F_{\theta}$ with learnable parameters $\theta$. 		

%In the CL based method in \citep{pmlr-v235-huang24f}, the set of positive samples ${S_p^{M_i}}$ is obtained by running state-of-the-art MILP solvers for a large amount of time. The set of negative samples ${S_n^{M_i}}$ is obtained by solving another MILP that searches within a neighborhood of the positive sample solution to find solutions with bad objective values (NEED TO REPHRASE).
%\citep{nair2020solving} and \citep{han2023a} assume conditional independence between variables. Given an instance $M$, let $x_d$ denote the $d^{th}$ element of a solution vector $x$. Conditional independence leads to $P_\theta(x;M)= \prod^n_{d=1}p_{\theta}\left(x_d;M\right)$. 
\paragraph{Inference}
Since the full prediction might not be feasible, ML-guided primal heuristics for MILPs involve solving another MILP at inference time. \citet{nair2020solving} use Neural Diving (ND), which uses the prediction of a subset of the variables and creates a smaller sub-MILP that is easier to solve after fixing the subset. The size of this sub-MILP is controlled by the ratio of variables that are fixed. \citet{han2023a} and \citet{pmlr-v235-huang24f} use a Predict-and-Search (PaS) framework that searches for feasible solutions within some neighborhood of the full prediction by adding a cut to the original MILP. The degrees of freedom in PaS are controlled by the number of variables that are allowed to be different from the prediction. The ND approach allows for faster runtime at inference time as the subproblem contains a small number of variables, but the solutions returned can be more suboptimal. PaS has more freedom to correct errors from the ML predictions, but it can be harder to optimize because the size of the MILP at inference contains the same number of variables as the original MILP.

\subsection{ML-Guided Methods for Nonlinear Optimization}
There has also been research on ML for general nonlinear optimization. Some of the works in this space focus on BnB and are not directly relevant to primal heuristics. For example, \citet{ghaddar2023learning} and \citet{maisonneuve2024learning} learn branching policies in BnB; \citet{baltean2019scoring} learns to select cutting planes; \citet{bonami2018learning} learns whether to linearize quadratic terms in the formulation before solving. Other works are relevant to primal heuristics, but do not apply to general MBQPs. \citet{saravanos2025deep} develops an ML-aided distributed optimization technique tailored to decomposable problems. \citet{liang2024generative} and \citet{gao2024ipmlstm} develop ML-guided methods for continuous problems, while \citet{bagga2023solving} develops an ML-based method tailored to the quadratic assignment problem. \citet{tang2025learningoptimizemixedintegernonlinear} develops a gradient-based method for inequality-constrained MINLPs using self-supervised learning and projection. To our knowledge, work on ML-guided methods for solving general MBQPs remains limited. The most closely related is \citet{ferber2023surco}, which develops SurCo, a gradient-based method for general MINLPs that can also be applied to MBQPs. Additionally, \citet{chen2024expressive} studies the theoretical expressiveness of graph neural networks for continuous and mixed-integer quadratic programs, which is complementary to our work. 

\section{Methods}\label{methods}
\begin{figure*}[h]
    \centering
    \includegraphics[width=0.8\textwidth]{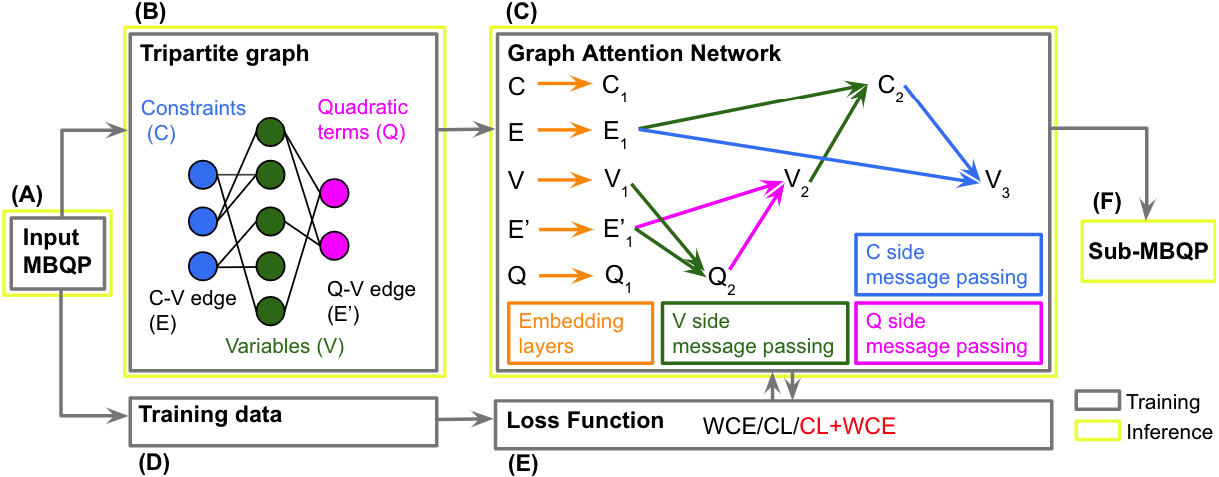}
    \caption{Training/Inference pipeline for ML-guided MBQP solving via solution prediction.}
    \label{fig:tri}
\end{figure*}

We develop an ML-guided primal heuristic for MBQPs based on solution prediction, as shown in Fig. \ref{fig:tri}. An input MBQP is represented as a tripartite graph (Fig. \ref{fig:tri} (B)) and then passed to a Graph Attention Network module (Fig. \ref{fig:tri} (C)), which produces solution predictions for binary decision variables in MBQPs. At inference time, the predicted solutions are used to create a sub-MBQP (Fig. \ref{fig:tri} (F)). We introduce a new method for collecting training data for MBQPs (Fig. \ref{fig:tri} (D)). In training the models, we adapt the WCE and CL losses which have been used in solution prediction in MILPs to MBQPs and propose an extended loss function that combines CL and WCE losses (Fig. \ref{fig:tri} (E)).

\subsection{Neural Network Architecture}
\paragraph{Tripartite Graph Representation} We propose a tripartite graph representation of MBQP instances (Fig. \ref{fig:tri} (B)). The tripartite graph contains three types of nodes: constraint nodes $(C)$, variable nodes $(V)$, and quadratic term nodes $(Q)$. A $C$--$V$ edge connects a variable node and a constraint node when the variable has a nonzero coefficient in that constraint. A $Q$--$V$ edge connects a quadratic term node to the two variable nodes that participate in the corresponding quadratic term. The feature sets for constraint and variable nodes are adapted from the MILP solution prediction framework of \citep{han2023a}. For quadratic term nodes, we introduce a custom feature set designed to capture the characteristics of quadratic terms in the MBQP objective function; the full list of features is provided in the appendix.

\paragraph{Graph Attention Network} We learn a policy $p(x|\theta;M)$ parameterized by $\theta$ that processes the featured tripartite graph and outputs predictions for the variables of a given instance $M$, using a Graph Attention Network (GAT)~\citep{brody2021attentive}. The GAT performs four rounds of message passing, as shown in Fig. \ref{fig:tri} (C). In round one, each quadratic term node in \(Q_1\) attends over its neighbors in $V_1$ using \(h\) attention heads to produce updated quadratic term embeddings \(Q_2\). In round two, each variable node in \(V_1\) attends over its neighbors in \(Q_2\) to produce updated variable embeddings \(V_2\). In round three, each constraint node in \(C_1\) attends over its neighbors in $V_2$ to produce updated constraint embeddings \(C_2\). In the final round, each variable node in \(V_2\) attends over its neighbors in \(C_2\) to produce the final variable embeddings \(V_3\). The message passing outputs are then passed through Multi-Layer Perceptrons (MLPs) to produce logits $z$. The final prediction is obtained by taking the sigmoid activation function $p(x|\theta;M)=\sigma(z)$ for the WCE-based method and the tanh activation $p(\theta;M)=\tanh(z)$ for the CL-based method. 

\subsection{Loss Function}
The solution prediction policy $p(x|\theta;M)$ can be learned with different approaches (Fig. \ref{fig:tri} (E)). In this work, we first adapt the WCE and CL losses that have been used for MILPs to MBQPs. Then, we propose an extension that combines CL and WCE losses to improve the performance. 
\paragraph{Weighted Cross-Entropy}
Following the Weighted Cross-Entropy (WCE) \citep{han2023a} approach, we create a training dataset that contains $N$ MBQP instances $\left\{\left(M_i,\;\mathcal{S}^+_{M_i}\right)\right\}^N_{i=1}$, where $\mathcal{S}^{+}_{M_i}$ is a set of unique solutions for the instance $M_i$. 
Let $p\left(x^+|\theta;M_i\right)$ denote the probability of solution $x^+$ given instance $M_i$ as the input. We adapt the energy function $E(x;M)$ in Eqn. (\ref{eq:predict:energy}) to the case of MBQPs to assign higher probability for better solutions. For a solution $x$, the energy function $E(x;M)$ is defined as $x^T H x + c^T x$ if $x$ is feasible. During training, for an instance $M_i$ with quadratic term matrix $H_i$ and cost vector $c_i$, the weight applied to the solution $x^+$ is 
$
			 w(x^+|M_i) \equiv \frac{\text{exp}\left(-{x^{+\top} H_i x^+}-{c_i}^{\top} x^+\right)}{\sum_{{x} \in \mathcal{S}^{+}_{M_i}}\exp{ \left(-x^{\top} H_i x-{c_i}^{\top}x\right)}} $
. Based on the Kullback-Leibler divergence, which measures the distance between the conditional distribution in Eqn. (\ref{eq:predict:energy}) and the learned policy, the loss function to be minimized is:
\begin{equation}
		\begin{aligned}
			\mathcal{L}^{\text{WCE}}\left(\theta\right) \equiv - \sum^N_{i=1} \sum_{{x}^{+} \in \mathcal{S}^{+}_{M_i}} w(x^+|M_i)\text{log} \;p\left(x^+| \theta;M_i\right)\\
		\end{aligned}\label{eq:kl}.
\end{equation}
To avoid numerical overflow, we follow prior work on WCE-based MILP solution prediction \citep{han2023a} and divide the objective value by a constant $\lambda$ (which is known as the weight-norm parameter) before computing the exponentials. Details on the weight-norm parameter are deferred to the appendix. 

\paragraph{Contrastive Learning}
Following the CL-based approach \citep{pmlr-v235-huang24f}, let $\left\{\left(\mathcal{S}^{+}_{M_i},\mathcal{S}^{-}_{M_i}\right)\right\}^N_{i=1}$ be a training dataset of $N$ MBQP instances, where $\mathcal{S}^{+}_{M_i}$ and $\mathcal{S}^{-}_{M_i}$ are the sets of positive and negative samples for instance ${M_i}$, respectively. We use a form of the NT-Xent Loss \citep{chen2020simple} to learn to distinguish between positive and negative samples. We use the $\cdot$ operator to denote the dot-product similarity. The loss function to be minimized is 
\begin{equation}
\mathcal{L}^{\text{CL}}({\theta})=\sum_{i=1}^N \sum_{x^+ \in \mathcal{S}^{+}_{M_i}} \mathcal{L}^+\left({\theta} \mid x^+, {M_i}\right)\label{eq:cl_tot},
\end{equation}
where
\begin{equation}
%\mathcal{L}^+\left({\theta} \mid x^+, {M_i}\right)= - \log \frac{\exp \left(x^+ \cdot p(\theta;{M_i}) / \tau(x^+|M_i)\right)}{\sum_{
%\Tilde{x} \in \mathcal{S}^{-}_{M_i} \cup \{x^+\}} \exp \left({\Tilde{x}}\cdot p(\theta;{M_i}) \right/ %\tau(\Tilde{x}|M_i))}.
\begin{split}
&\mathcal{L}^+\left({\theta} \mid x^+, {M_i}\right)\\
&= - \log \frac{\exp \left(x^+ \cdot p(\theta;M_i) / \tau(x^+|M_i)\right)}{\sum_{
\Tilde{x} \in \mathcal{S}^{-}_{M_i}\cup \{x^+\}} \exp \left({\Tilde{x}}\cdot p(\theta;M_i) \right/\tau(x^+|M_i) )}.
\end{split}
\label{eq:cl}
\end{equation}

%The NT-Xent Loss measures similarity with dot products and takes into account the solution qualities of both positive and negative samples.
We use a -1/1 encoding for the solution values in order to measure similarity when computing the dot-product in Eqn. (\ref{eq:cl}). Solution values of 0 are encoded as -1, while solution values of 1 remain 1 in $x^+$ and $\Tilde{x}$. To match the -1/1 encoding, we convert the prediction to be in the range of $[-1,1]$ by taking the tanh activation function for the logits, (i.e., $p(\theta;M_i)=\tanh(z_i)$ where $z_i$ is the logits obtained given instance $M_i$ as the input). Based on the dot-product similarity, the loss value $\mathcal{L}^+\left({\theta} \mid x^+, {M_i}\right)$ is low when $p(\theta;M_i)$ is similar to the positive sample $x^+$ and dissimilar to negative samples $\Tilde{x} \in \mathcal{S}^{-}_{M_i} $. $\tau(x|M_i)$ is a temperature parameter that scales the similarity scores in Eqn. (\ref{eq:cl}). We set 
${\tau(x|{M_i})}= \frac{1}{\exp((x^T H_i x + c_i^T x)/{s})}$
with $s<0$ for minimization problems, so that the loss $\mathcal{L}^+\left({\theta} \mid x^+, {M_i}\right)$ for better $x^+$ (i.e., positive sample with a lower objective value) has a lower temperature parameter, which creates a sharper distribution and increases the penalties on the negative samples. 
%Notes:

%\paragraph{Loss actually used in Neurips workshop workshop submission (MBQP) and CL-PAS (MILP):}
%\begin{equation}
%\mathcal{L}^+\left({\theta} \mid x^+, {M_i}\right)= - \log \frac{\exp \left(x^+ \cdot p(\theta;{M_i}) / \tau(x^+|M_i)\right)}{\sum_{
%\Tilde{x} \in \mathcal{S}^{-}_{M_i}} \exp \left({\Tilde{x}}\cdot p(\theta;{M_i}) \right )}.\label{eq:cl2}
%\end{equation}

%\paragraph{New proposed:}
%\begin{equation}
%\mathcal{L}^+\left({\theta} \mid x^+, {M_i}\right)= - \log \frac{\exp \left(x^+ \cdot p(\theta;{M_i}) / \tau(x^+|M_i)\right)}{\sum_{
%\Tilde{x} \in \mathcal{S}^{-}_{M_i}} \exp \left({\Tilde{x}}\cdot p(\theta;{M_i}) \right/\tau(x^+|M_i) )}.\label{eq:cl3}
%\end{equation}

\paragraph{Combining Contrastive Learning and Weighted Cross-Entropy}\label{comb}
In addition to extending loss functions used in MILP solution prediction, we propose to combine CL and WCE losses. It has been observed that for each positive sample $x^+$, a subset of variables often have the same assignments across $x^+$ and the corresponding negative samples in the CL-based approach, and that the CL loss $\mathcal{L}^+\left({\theta} \mid x^+, {M_i}\right)$ in Eqn. (\ref{eq:cl}) does not depend on the predicted solution by the ML model for this subset. This observation is formalized in Proposition \ref{prop:1}; the proof is deferred to the appendix. 
%We also show empirically in Section. \ref{results} that the percentage of this subset can be large in MBQP benchmarks.

\begin{prop}
	Given an MBQP instance $M_i$ with sets of positive and negative samples $\left(\mathcal{S}^{+}_{M_i},\mathcal{S}^{-}_{M_i}\right)$, let $\mathcal{B}_i$ be the index set of all binary decision variables, and let $x^+ \in \mathcal{S}^{+}_{M_i}$ be any positive sample. Let $\mathcal{U}_{i}^{x^{+}} \subseteq \mathcal{B}_i$ be an index set on which the positive sample and all corresponding negative samples agree; that is, $\tilde{x}_d = t_d \; \forall d \in  \mathcal{U}_{i}^{x^{+}} \ \forall \tilde{x} \in \mathcal{S}^-_{M_i} \cup \{x^+\}$ where $t_d \in \{0,1\}$ is the common assignment of the variable indexed by $d$. The CL loss $\mathcal{L}^+\left({\theta} \mid x^+, {M_i}\right)$ in Eqn. (\ref{eq:cl}) depends only on the predictions for the subset of variables indexed by  $\mathcal{B}_i\setminus {\mathcal{U}^{x^+}_i}$.
    
    %Let $x_d$ denote the assignment for the $k^{th}$ variable in solution $x$..$\forall d \in {\mathcal{U}_{i}^{x^{+}}}$ it holds that $\exists t_d^i \in \{0,1\} \text{ s.t. } x_d^i=t_d^i, \forall x \in \mathcal{S}^{-}_{M_i} \cup \{x^+\}$. 

	\label{prop:1}
\end{prop}
As shown in Proposition~\ref{prop:1}, for each positive sample $x^+$, the CL loss $\mathcal{L}^+\left(\theta \mid x^+, M_i\right)$ penalizes predictions that resemble negative samples only for variables outside the subset $\mathcal{U}^{x^+}_i$. Consequently, it provides no gradient signal for variables in ${\mathcal{U}^{x^+}_i}$. To improve the accuracy of the ML model in predicting the solution for variables in ${\mathcal{U}^{x^+}_i}$, we propose to apply classification of whether the variable takes 1 or 0 as the solution value using binary Cross-Entropy (CE) loss, given that variables in this set take the same solution value in $\mathcal{S}^{-}_{M_i} \cup \{x^+\}$ by definition. For each positive sample $x^+$, we apply the CL loss in Eqn. (\ref{eq:cl}) (which only applies to $\mathcal{B}_i\setminus {\mathcal{U}^{x^+}_i}$) and CE loss for ${\mathcal{U}^{x^+}_i}$. For each instance $M_i$, the CE loss for each $x^+ \in \mathcal{S}^{+}_{M_i}$ is weighted by $w(x^+|M_i)$, the weight derived from the adapted energy function in Eqn. (\ref{eq:predict:energy}). Formally, for each positive sample $x^+ \in \mathcal{S}^{+}_{M_i}$, let $t_{d,i}$ denote the assignment of the $d$-th variable, which is the same across $\mathcal{S}^{-}_{M_i} \cup \{x^+\}$ for $d \in \mathcal{U}^{x^+}_i$. Let $\hat p_{d,i} \equiv p\left(x_{d,i}=1|\theta; M_i\right)$ be the probability that the $d$-th variable of $M_i$ takes a solution value of 1 predicted by the ML policy. The CE loss for each $x^+$ is $\mathcal{L}^{\text{CE}}({\theta} \mid  x^+, {M_i})=-\sum_{d \in \mathcal{U}^{x^+}_i}\left[ t_{d,i} \log(\hat p_{d,i}) + (1 - t_{d,i}) \log(1 - \hat p_{d,i})\right]$. Accounting for the weights for each positive sample, the combined loss function to be minimized is 
\begin{equation}
\begin{split}
\mathcal{L}^{\text{CL+WCE}}(\theta) &= \sum_{i=1}^{N} \sum_{x^+ \in \mathcal{S}^{+}_{M_i}} \Big( \lambda^{\text{CL}}\, \mathcal{L}^+\left(\theta \mid x^+, M_i\right) \\
&+ w(x^+ \mid M_i)\, \mathcal{L}^{\text{CE}}(\theta \mid x^+, M_i) \Big), \label{eq:new_comn}
\end{split}
\end{equation}
where 
$\lambda^{CL}$ is a hyperparameter that controls the weight of the CL loss compared to WCE loss. 

\subsection{Training Data Collection for MBQPs}\label{rand_rs}
Training data collection (Fig. \ref{fig:tri}(D)) aims to compile multiple high-quality solutions, as well as low-quality solutions for the CL-based approach, for use in MBQP solution prediction. To this end, we propose \emph{Randomized Relax-Search}, a novel heuristic that generates a diverse set of high-quality solutions for MBQPs. \emph{Randomized Relax-Search} extends \emph{Relax-Search} \citep{Huangsocs_2025}, which constructs a subproblem from a suboptimal relaxation solution of the MBQP. As shown in Algorithm \ref{alg:randomized_greedy}, \emph{Randomized Relax-Search} first forms a candidate set $C$ consisting of variables that are least fractional in the relaxation of $\mathcal{P}$, and then introduces randomization to construct $K$ subproblems of $\mathcal{P}$. To construct the $k^{th}$ subproblem, a subset $C'_k \subset C$ is selected uniformly at random, and constraints are added to $\mathcal{P}$ to fix each variable $i \in C'_k$ to its nearest integer value. In solving the $k^{th}$ sub-MBQP, the best solution $x_k^+$ and the worst solution $x_k^-$ are stored. The procedure returns sets of best solutions $S^+$ and worst solutions $S^-$ of $\mathcal{P}$.

\begin{algorithm}[t]
    \caption{Randomized Relax-Search for training data collection}
    \label{alg:randomized_greedy}
    \begin{flushleft}
        \textbf{Input:} An MBQP $\mathcal{P}$ with set of binary variables $\mathcal{B}$, relaxation time limit $T_r$, subproblem time limit $T_s$, number of random seeds $K$, candidate fixing ratio $p_1$, final fixing ratio $p_2$ ($p_1>p_2$) \\
        \textbf{Output:} Set of good solutions $S^+$, Set of bad solutions $S^-$
\end{flushleft}
    \begin{algorithmic}[1]
       
        %, fixing ratio $p$
        %\Ensure An approximate solution to [Your Problem Name], e.g., a subset $R \subseteq S$.

        \State Relaxed solution $\bar{x} \gets$ Compute the Nonlinear Programming relaxation of $\mathcal{P}$ given time limit $T_r$
        \State $S^+, S^- \gets \emptyset$
        \State Candidate set $\mathcal{C} \gets$ select $p_1*|\mathcal{B}|$ variables that are least fractional in $\bar{x}$
        \For { $k \in {1,2,...,K}$}
            \State $\mathcal{C'}_k$  $\gets$ Uniformly at random select $p_2*|\mathcal{B}|$ variables from $\mathcal{C}$
            \State $\mathcal{P}_k$  $\gets$  $\mathcal{P} \cup \{x_i = \lfloor\bar{x_i}\rceil\}_{i \in \mathcal{C'}_k} $ 
            \State  $x_k^+,x_k^-  \gets$ Best and worst solutions obtained by solving subproblem $\mathcal{P}_k$ with a complete solver, given time limit $T_s$
            \State $S^+ \gets S^+\cup \{x_k^+\}$
            \State $S^- \gets S^-\cup \{x_k^-\}$

        \EndFor
        \State \Return $S^+,S^-$
    \end{algorithmic}
\end{algorithm}

\begin{table*}[h]

\centering
\scalebox{0.90}{
\begin{tabular}{llllll}
\toprule
Benchmark & $|\mathcal{S}^{+}|_{\text{SCIP}}$ & $|\mathcal{S}^+|_{\text{Rand-Relax-Search}}$ &  obj. $\mathcal{S}^{+}_{\text{SCIP}}$  & obj. $\mathcal{S}^{+}_{\text{Rand-Relax-Search}}$ & \text{frac.} $\mathcal{U}$\\
\midrule
CBQP       & 2.01          & 10          & -289048.26                                     & -887895.29                    & 59.71\%           \\
QMKP       & 4.31           & 10           & -46736.29                                    & -177384.21                    & 72.48\%              \\
CQKP      &       2.47            & 10        &        -58559.04                                      & -187403.44                        & 56.74\%         \\
WFLOP      &        8.59             & 10       &          1999.23                                   & 1478.76      & 57.06\%  \\              
\bottomrule
\end{tabular}}
\caption{
Data collection statistics with proposed Randomized Relax-Search vs SCIP. Average number of distinct solutions found by SCIP ($|\mathcal{S}^{+}|_{\text{SCIP}}$), average number of distinct solutions in the good samples set $\mathcal{S}^{+}$ returned by Randomized Relax-Search ($|\mathcal{S}^+|_{\text{Rand-Relax-Search}}$), average objective value found by SCIP (obj. $\mathcal{S}^{+}_{\text{SCIP}}$), average objective value in $\mathcal{S}^{+}$ by Randomized Relax-Search (obj. $\mathcal{S}^{+}_{\text{Rand-Relax-Search}}$), and average percentage of variables that take the same value in positive and negative sample pairs ($\text{frac. }\mathcal{U}$). The time limit for both strategies is 11000s.  
}
\label{tab:collect_stats}
\end{table*}
For training with WCE loss, the set of good solutions $S^+$ is used. For CL losses, $S^+$ is used as the set of positive samples. We denote the worst solution value from $S^+$ as $v'= \max_{x \in S^+} x^THx+c^Tx$. For the set of negative samples in CL, we use $\{x|(x^THx+c^Tx)>v',x \in S^-\}$. In other words, we only include solutions in $S^-$ that have worse objective values than the worst solutions in $S^+$.

\subsection{Inference}
At inference time, we choose to use the ND-based method discussed in the Background and Related Works section, which reduces the original problem to a smaller sub-MBQP (Fig. \ref{fig:tri} (F)), as our goal is to develop fast primal heuristics. The PaS-based method is challenging for MBQPs because it requires solving another MBQP of the same size (number of binary variables). After obtaining the variable predictions, we create a sub-MBQP by fixing the top $p$ percent of variables,  for which the ML model is most confident (i.e., least fractional in the predictions). The resulting sub-MBQP is then solved with a CO solver.

\section{Computational Experiments}\label{experiemtns}

\begin{table*}[ht]

\centering
\scalebox{0.85}{
\setlength{\tabcolsep}{5pt}
\begin{tabular}{llllllllllllll}
\toprule
                &            & \multicolumn{3}{c}{\bfseries CBQP} & \multicolumn{3}{c}{\bfseries QMKP} & \multicolumn{3}{c}{\bfseries CQKP} & \multicolumn{3}{c}{\bfseries WFLOP} \\
& \bfseries Method  & \bfseries PG & \bfseries PI &  \bfseries \# wins & \bfseries PG & \bfseries PI & \bfseries \# wins & \bfseries PG & \bfseries PI &  \bfseries \# wins & \bfseries PG  & \bfseries PI & \bfseries \# wins   \\
                              \midrule
Adapted & WCE          & 0.12                   & 33.21                  & 20                             & 0.09                   & 24.98                  & 26                             & 0.16                   & 36.32                  & 14                             & 0.27                   & 34.59                  & 23   \\
& CL   & 0.3                    & 40.69                  & 6                              & 0.98$^\dagger$                   & 59.21$^\dagger$                  & 0$^\dagger$                              & 1$^\dagger$                      & 60$^\dagger$                     & 0$^\dagger$                              & 0.83$^\dagger$                   & 53.46$^\dagger$                  & 7$^\dagger$  \\
 \midrule
Extended  & CL+WCE   & \textbf{0.02}          & \textbf{22.4}          & \textbf{74}                    & \textbf{0.04}          & \textbf{21.54}         & \textbf{74}                    & \textbf{0.06}          & \textbf{27.07}         & \textbf{82}                    & \textbf{0.14}          & \textbf{30.96}         & \textbf{40}   \\

%BCE+CL ($\lambda_{CL}=5$)   & \bf{0.03}   & \bf{46.73}        & 0.54       & 172.57       & 0.07    & \bf{51.79}           & 0.06       & 67.28          \\
 \midrule
 \midrule
Baselines & SCIP                 & 1                      & 60                     & 0                              & 0.89                   & 57.83                  & 0                              & 1                      & 60                     & 0                              & 0.59                    & 39.52                   & 11    \\
& RENS            & 1$^\dagger$                      & 60$^\dagger$                     & 0$^\dagger$                              & 1                      & 59.96                  & 0                              & 0.99$^\dagger$                   & 59.69$^\dagger$                  & 0$^\dagger$                              & 0.36                    & 39.45                & 18    \\
& Undercover          & 1$^\dagger$                      & 60$^\dagger$                     & 0$^\dagger$                              & 1                      & 60                     & 0                              & 1                      & 60                     & 0                              & 0.99                   & 59.73                  & 0    \\
& Relax-Search     & 0.57                   & 50.49                  & 0                              & 0.62                   & 52.31                  & 0                              & 0.56                   & 52.32                  & 4                              & 0.54                   & 48.9                 & 1   \\ 
& SurCo      & 1                      & 60                     & 0                              & 1                      & 60                     & 0                              & 1                      & 60                     & 0                              & 0.59                   & 49.06                  & 0   \\ 
\bottomrule 
\end{tabular}
}
\caption{
Primal Gap (PG), Primal Integral (PI), and \# wins in terms of PI. Lower is better for PG and PI. Higher is better for \# wins. WCE and CL are ML-guided MBQP primal heuristics that we adapted from MILPs. CL+WCE is the extended ML method with the proposed combined loss function. $^\dagger$ indicates benchmarks where there are instances for which the method did not produce a feasible solution. For CL, the feasibility rates are 2\%, 0\%, and 23\% for QMKP, CQKP, and WFLOP. For RENS, the feasibility rates are 0\% and 14\% for CBQP and CQKP. For Undercover, the feasibility rate is 99\% for CBQP. For all other methods and benchmarks, the feasibility rate is 100\%. Note that PG, PI, and \# wins apply to all instances regardless of feasibility.
}\label{tab:res}
\end{table*}
\begin{figure*}[ht]%!htbp
   \centering
   %\begin{subfigure}[htbp]{0.8\textwidth}
   \includegraphics[width=0.8\textwidth]{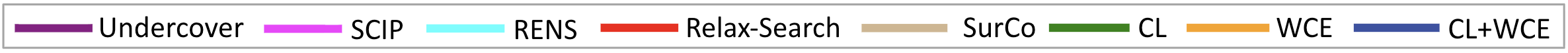}
%\end{subfigure}

    \begin{subfigure}[htbp]{0.24\textwidth}
		\centering
		\includegraphics[width=0.99\textwidth]{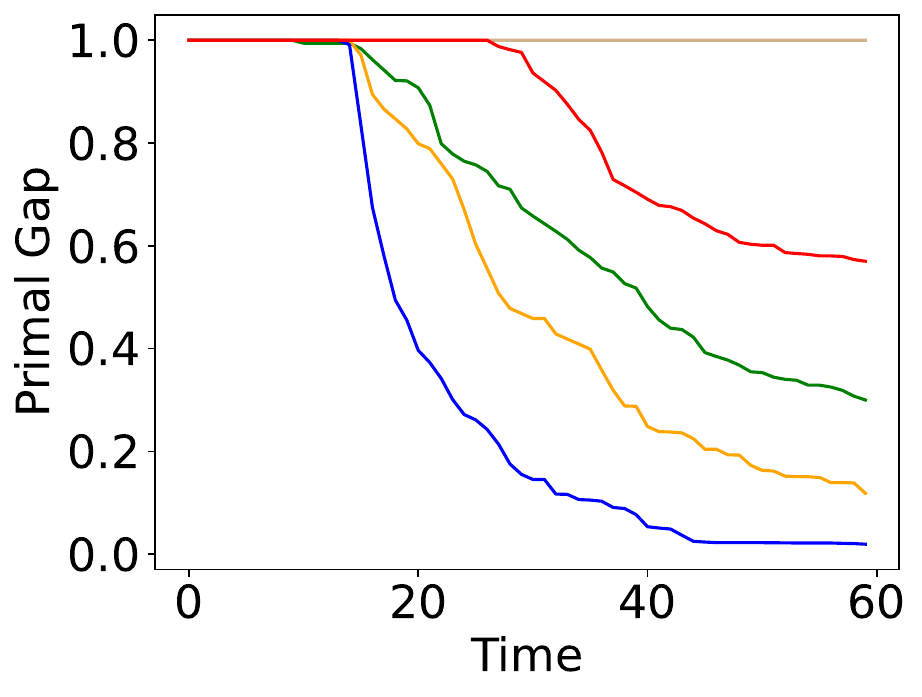}
		\caption{CBQP}\label{fig:1a}
		%\caption{An example with one agent.}
	\end{subfigure}
    \hspace{0em}
	\begin{subfigure}[htbp]{0.24\textwidth}
		\centering
		\includegraphics[width=0.99\textwidth]{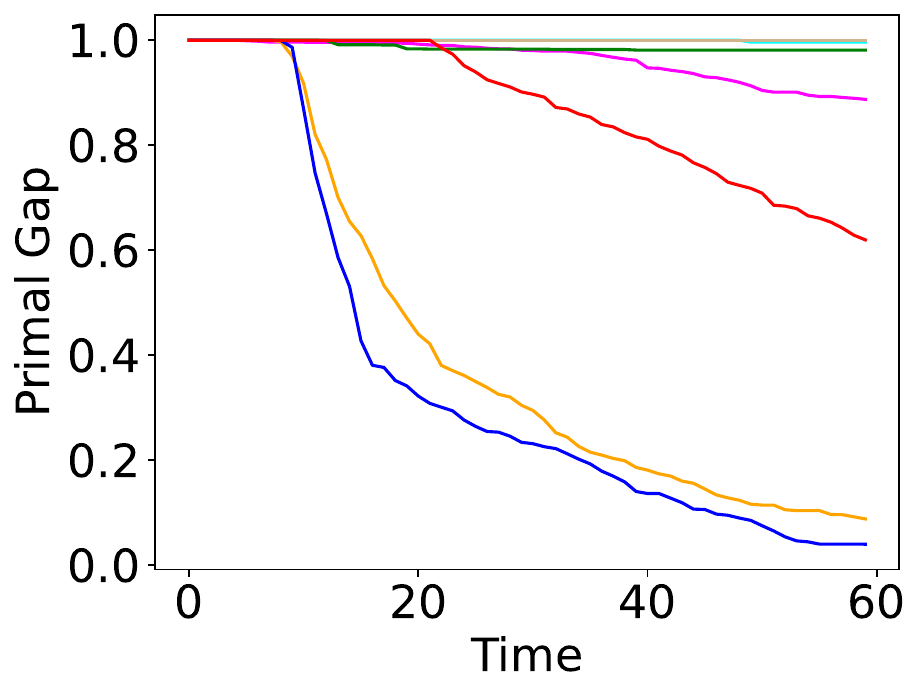}
			\caption{QMKP}
	\end{subfigure}
    \hspace{0em}
	\begin{subfigure}[htbp]{0.24\textwidth}
		\centering
		\includegraphics[width=0.99\textwidth]{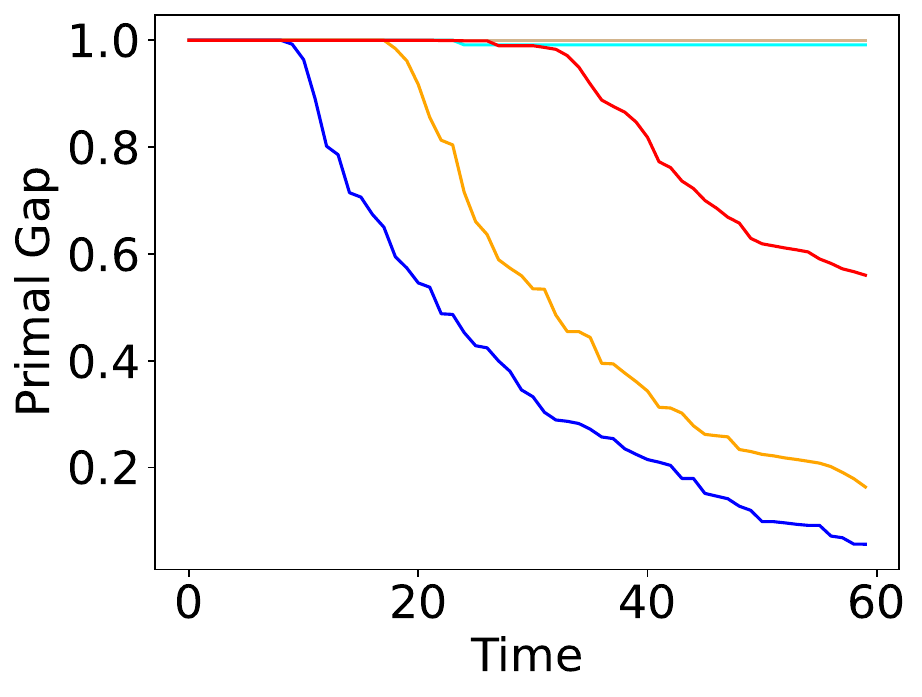}
			\caption{CQKP}
	\end{subfigure}
    \hspace{0em}
	\begin{subfigure}[htbp]{0.24\textwidth}
		\centering
		\includegraphics[width=0.99\textwidth]{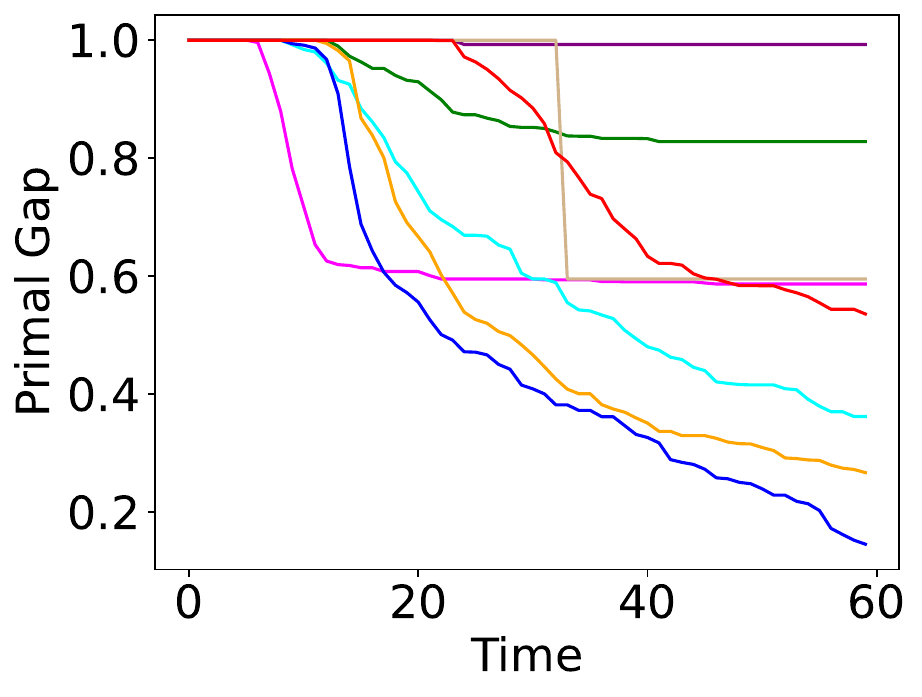}
			\caption{WFLOP}
	\end{subfigure}
    \caption{Primal gap as a function of time (lower is better).
    \label{fig_primal_gap}}
\end{figure*}

\subsection{Setup}\label{comp_Setup}

\paragraph{Benchmarks} We evaluate the methods on synthetic and real-world benchmarks. For synthetic benchmarks, we include the Cardinality-constrained Binary Quadratic Programs (CBQP) \citep{zheng2012successive}, Cardinality-constrained Quadratic Knapsack Problem (CQKP) \citep{letocart2014efficient}, and the Quadratic Multidimensional Knapsack Problem (QMKP) \citep{forrester2020computational}. All synthetic benchmark instances contain 1000 binary variables and have a quadratic term density of $25\%$. Moreover, we test on a real-world \emph{Wind Farm Layout Optimization Problem} (WFLOP). WFLOP seeks to identify the placement of a set of wind turbines within a fixed area to maximize power generation across all turbines and over all wind scenarios while also satisfying minimum separation constraints. We use the MBQP formulation of WFLOP in \citep{Huangsocs_2025} and instantiate the instances using probabilistic wind models (i.e., probability density functions sampled from the NOW-23 dataset \citep{bodini20232023}) which represents long-term wind patterns over a 10 year time horizon at selected locations in the California offshore region. The WFLOP instances contain 1000 binary variables while the quadratic term densities depend on the wind distribution. On average, the WFLOP-California instances have a quadratic term density of $32.42\%$. Details on the benchmarks are deferred to the appendix.

%We follow the approach described in \citep{forrester2020computational} to generate random $H$ matrices for general nonconvex MBQPs. 

\paragraph{Evaluation Metrics} We use the following metrics to evaluate the effectiveness
of different methods:
(1) The \textit{Primal Gap} (PG) \citep{berthold2013measuring} is the normalized difference between the objective value $v$ found by a method and a best known objective value $v^*$, defined as 
    $\text{PG} = \frac{|v-v^*|}{max(|v|,|v^*|)}$, when $vv^*>0$. When no feasible solution is found or when $vv^* <0$, PG is defined to be 1. PG is 0 when $|v|=|v^*|=0$. (2) The \textit{Primal Integral} (PI) \citep{berthold2013measuring} is the integral of the primal gap over time, which captures the speed at which better solutions are found. (3) The \textit{\# wins} in terms of PI is the number of test instances for which the method results in the lowest PI across all other methods.

\paragraph{Baselines}
We compare the proposed ML-guided MBQP solving methods adapted from MILPs (WCE and CL) and the proposed method with the extended loss function (CL+WCE) with well-established primal heuristics for MBQPs and general MINLPs, including RENS \citep{berthold2014rens}, Undercover \citep{berthold2014undercover}, and Relax-Search \citep{Huangsocs_2025}. In addition, we compare with the state-of-the-art MINLP solver SCIP \citep{bolusani2024scip}, which uses BnB as its core component and includes primal heuristics as supplementary procedures to improve the primal bound during BnB. We turn on the aggressive mode in SCIP to focus on finding good primal solutions instead of improving the dual bound. We experiment with both the Linear Programming (LP) reformulation (default) and Nonlinear Programming (NLP) formulation in SCIP and report the stronger one. To our knowledge, this is the first work on ML-guided primal heuristics tailored to general MBQPs. As an ML-guided baseline, we include SurCo \cite{ferber2023surco}, an ML-based primal heuristic for general MINLPs.

\paragraph{Hyperparameters}
For the proposed combined loss, we experiment with $\lambda^{CL} \in \{1,2,5,7\}$ and choose the $\lambda^{CL}$ value that results in the lowest weighted Brier Score (BS) on the validation set. Specifically, BS is defined as 
\[
BS = \frac{1}{N} \sum_{i=1}^{N} \sum_{x^+ \in {{\mathcal{S}^{+}_{M_i}}}} \sum_{d\in \mathcal{B}_i}w(x^+|M_i)(\hat p_{d,i} - t^+_{d,i})^2
\]
where $t^+_{d,i}$ is the solution assignment of the variable indexed by $d$ in sample $x^+$ from instance $M_i$. $\mathcal{B}_i$ is the set of binary variables in $M_i$. $N$ is the total number of instances in the validation set. For the hyperparameter $p$, which specifies the fraction of variables fixed when constructing the sub-MBQP at inference time, we experiment with $p \in \{0.5,0.6,0.7,0.8,0.9\}$ on the validation set and choose the best $p$.
%The value of $\lambda^{CL}$ is 1, 2, 5, and 2 for CBQP, QMKP, CQKP, and WFLOP-California. $\lambda^{CL}$ is 7 for WFLOP-mix.

\paragraph{Computational Setup}
For all methods, we set the time limit to 60s. We report the average PG, PI, and \# wins results across 100 test instances for all benchmarks. All models are trained on 800 instances, and training is conducted on an NVIDIA A100 GPU on a node with 128 GB of RAM. We use the AdamW optimizer \citep{loshchilov2017decoupled} with learning rate $10^{-5}$ with a batch size of 16. Testing (including ML inference and non-ML primal heuristics) is conducted on a cluster with epyc-7542 CPUs with 10 GB RAM. We use SCIP (v9.0) \citep{bolusani2024scip} for solving the sub-MBQPs.

\begin{table*}[ht]

\centering
\scalebox{0.85}{
\setlength{\tabcolsep}{4pt}
\begin{tabular}{lllllllllll}
\toprule
                &            & \multicolumn{3}{c}{\bfseries California} & \multicolumn{3}{c}{\bfseries Hawaii} & \multicolumn{3}{c}{\bfseries Great Lakes}  \\
& \bfseries Method  & \bfseries PG & \bfseries PI &  \bfseries \# wins & \bfseries PG & \bfseries PI & \bfseries \# wins & \bfseries PG & \bfseries PI &  \bfseries \# wins   \\
                              \midrule
Adapted   & WCE & 0.29                   & 36.78                  & 12                             & 0.31                   & 42.07                  & 9                              & \textbf{0.19}          & 40.16                  & 10                 \\
& CL     & 0.71$^\dagger$                   & 48.57$^\dagger$                  & 17$^\dagger$                             & 0.71$^\dagger$                   & 49.97$^\dagger$                  & 24$^\dagger$                             & 0.88$^\dagger$                   & 55.92$^\dagger$                  & 9$^\dagger$  \\
 \midrule
Extended  & CL+WCE   & \textbf{0.24}          & \textbf{31.71}         & \textbf{34}                    & \textbf{0.29}          & 38.57                  & \textbf{25}                    & \textbf{0.19}          & \textbf{36.3}          & \textbf{46}     \\

 \midrule
 \midrule
Baselines & SCIP           & 0.58                   & 39.01                  & 8                              & 0.64                   & 41.83                  & 10                             & 0.61                   & 41.1                   & 3  \\
& RENS        & 0.32                   & 34.95                  & 20                             & 0.45                   & \textbf{38.47}         & 24                             & 0.38                   & 36.76                  & 31    \\
& Undercover         & 0.98                   & 59.28                  & 0                              & 1                      & 60                     & 0                              & 1                      & 60                     & 0    \\
& Relax-Search & 0.43                   & 38.64                  & 9                              & 0.61                   & 42.04                  & 9                              & 0.56                   & 42.58                  & 1   \\ 
& SurCo    & 0.59                   & 45.91                  & 0                              & 0.37                   & 46.78                  & 0                              & 0.3                    & 44.65                  & 0         \\ 
\bottomrule 
\end{tabular}
}
\caption{
Cross-regional generalization results on WFLOP. Primal Gap (PG), Primal Integral (PI) results, and \# wins in PI. Lower is better for PG and PI. Higher is better for \# wins. $^\dagger$ indicates benchmarks where there are instances for which the method did not produce a feasible solution. For CL, the feasibility rates are 36\%, 34\%, and 13\% for California, Hawaii, and Great Lakes. Note that PG, PI, and \# wins apply to all instances regardless of feasibility.
\vspace{0.1cm}
}\label{tab:corssresults}
\end{table*}

\begin{figure*}[ht]%!htbp
   \centering
   %\begin{subfigure}[htbp]{0.8\textwidth}
   \includegraphics[width=0.8\textwidth]{PG_plots/legend.png}
%\end{subfigure}

    \begin{subfigure}[htbp]{0.24\textwidth}
		\centering
		\includegraphics[width=0.99\textwidth]{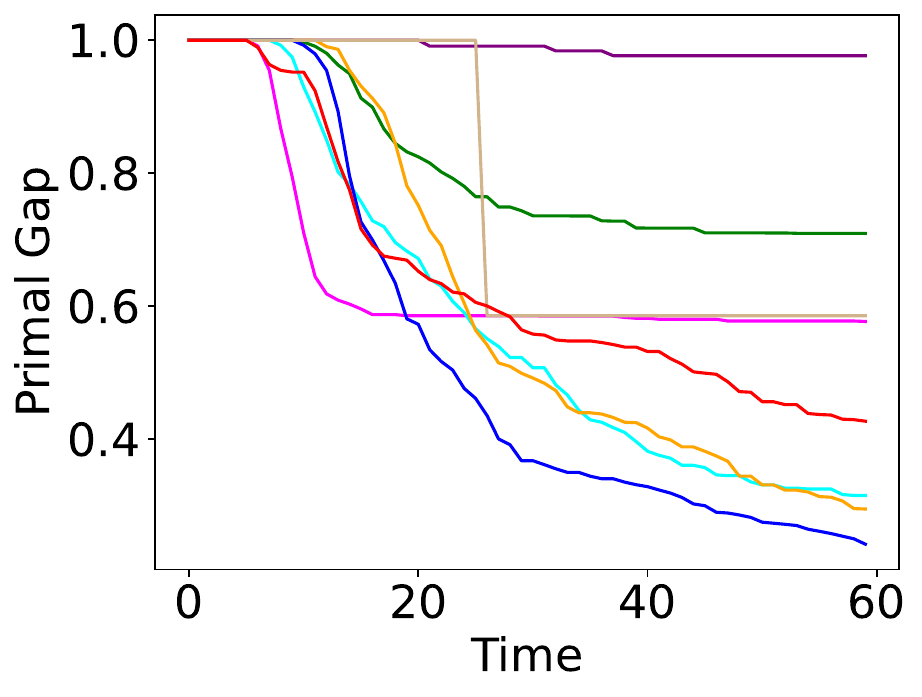}
		\caption{California}%\label{fig:1a_mix}
		%\caption{An example with one agent.}
	\end{subfigure}
    \hspace{0em}
	\begin{subfigure}[htbp]{0.24\textwidth}
		\centering
		\includegraphics[width=0.99\textwidth]{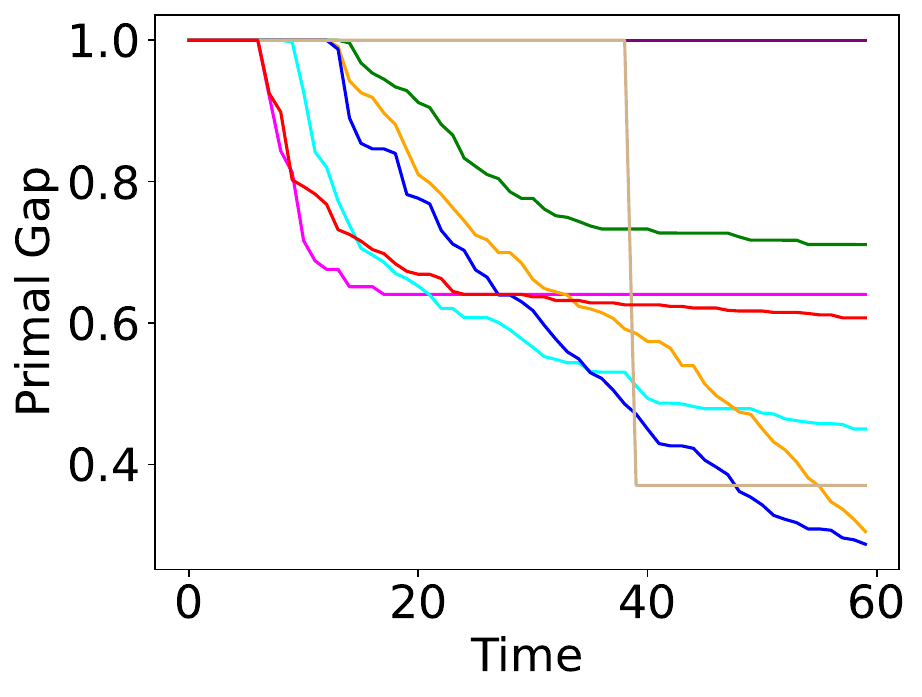}
			\caption{Hawaii}
	\end{subfigure}
    \hspace{0em}
	\begin{subfigure}[htbp]{0.24\textwidth}
		\centering
		\includegraphics[width=0.99\textwidth]{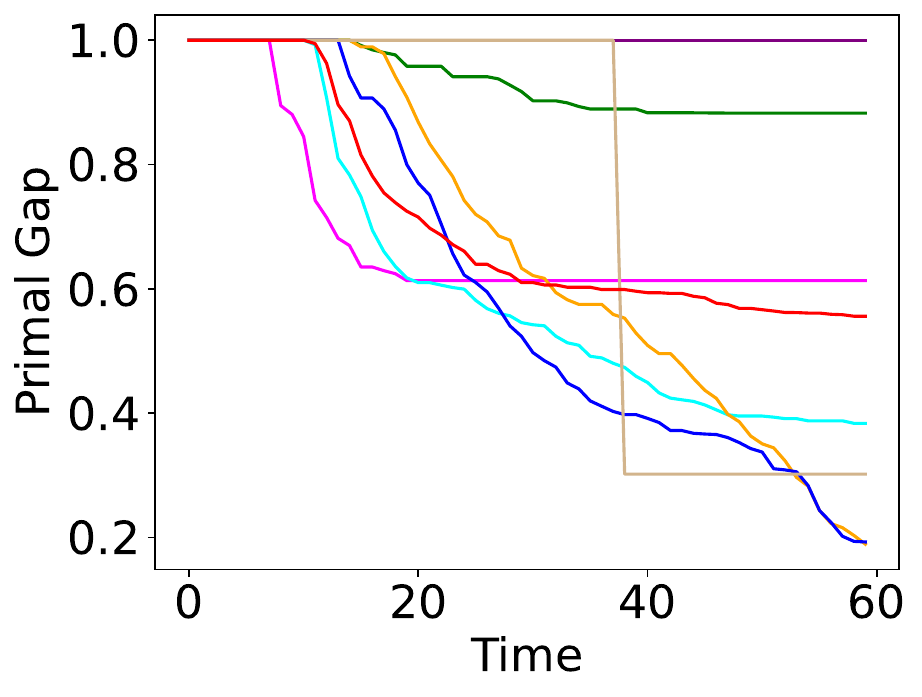}
			\caption{Great Lakes}
	\end{subfigure}

    \caption{Primal gap as a function of time (lower is better).
    \label{fig_primal_gap_tranfer}}
\end{figure*}
\paragraph{Training Data Collection} We compare the training data collection results with \emph{Randomized Relax-Search} and running the SCIP solver under the same time limit. As shown in Table \ref{tab:collect_stats}, the number of distinct solutions found by SCIP throughout the time limit is less than 10 across all benchmarks. Moreover, \emph{Randomized Relax-Search} produces solutions with better (i.e., lower for minimization problems) objective values in the positive sample set $S^+$. Additionally, we report the average fraction of the subset of variables that take the same value in positive and negative sample pairs discussed in Proposition \ref{prop:1}, $\text{frac. }\mathcal{U}=\frac{1}{N}\sum_{i=1}^N \frac{1}{|\mathcal{S}^{+}_{M_i}|}\sum_{x^+ \in \mathcal{S}^{+}_{M_i}}\frac{|\mathcal{U}^{x^+}_i|}{|\mathcal{B}_i|}$.

%\begin{table}[h]
%\caption{
%\textbf{Primal Integral (PI) results at 300s time cutoff.} WCE and CL are adapted from ML-guided primal heuristics for MILPs. BCE+CL, $\lambda_{CL} \in \{1,2,5\}$ are the new proposed losses. SCIP, RENS, Undercover and Relax-Search are non-ML baselines. 
%\vspace{0.1cm}
%}\label{tab:res_PI}
%\centering
%\begin{tabular}{lllll}
%\toprule
%                            & {\bfseries CBQP} & {\bfseries QMKP} &{\bfseries CQMKP} & {\bfseries WFLOP} \\
%                              \midrule
%WCE             & \textbf{50.2}              & 78.22           & 68.38           & 66.28           \\
%CL & 109.62         & 294.77          & 300             & 274.94          \\
% \midrule
%BCE+CL ($\lambda_{CL}=1$)     & 52.53           & 67.16           & 74.84            & 65.6           \\
%BCE+CL ($\lambda_{CL}=2$)   & 52.25           & \bf{55.17}            & \textbf{62.13}             & \bf{64.98}            \\
%BCE+CL ($\lambda_{CL}=5$)   & \bf{46.73}            & 172.57           & \bf{51.79}           & 67.28         \\
% \midrule
% \midrule
%SCIP                     & 278.22           & 265.9             & 298.27           & 153.69           \\
%RENS                    & 276.02       & 282.59      & 292.86          & 116.25          \\
%Undercover                      & 300                & 299.79         & 298.36        & 262.56           \\
%Relax-Search            & 183.6             & 182.76           & 163.56          & 150.09      \\ 
%\bottomrule 
%\end{tabular}
%\end{table}

\paragraph{Inference}

As shown in Table \ref{tab:res}, both the adapted WCE method and the extended CL+WCE method outperform the baselines in terms of average PG and PI. The adapted CL-based method fails to produce feasible solutions for some instances in QMKP, CQKP, and WFLOP. The extended CL+WCE method performs best in PG, PI, and the number of PI wins across all four benchmarks, and it significantly improves both feasibility and solution quality over the adapted CL-based method. Fig. \ref{fig_primal_gap} shows the progress of PG over time. The proposed CL+WCE method finds better solutions at all time steps in most benchmarks (the exception being that SCIP achieves lower PG early on in WFLOP), with the adapted WCE method being the second-best. 

Proposition~\ref{prop:1} explains why CL+WCE outperforms CL. In the four benchmarks studied, over 50\% of the binary variables take the same assignment across the positive sample and its corresponding negative samples (Table~\ref{tab:collect_stats}). By Proposition~\ref{prop:1}, the CL loss provides zero gradient signal for these variables, which explains both why pure CL underperforms in PG and PI and why it has low feasibility rates. WCE, in contrast, supervises every variable but is less discriminative on the variables where positive and negative samples differ. CL+WCE combines the two: it applies CL to the variables where negatives differ from positives and WCE to those that share the same assignment across positive and negative samples. Combining the two losses therefore improves solution-prediction performance on this substantial subset of variables, which justifies our proposed extension.

%This fraction exceeds $50\%$ across all four MBQP benchmarks, which justifies our proposed extension of combining CL and WCE losses to improve the solution prediction performance for this subset. The reason why CL+WCE outperforms CL can be explained by Proposition \ref{prop:1}. In the four benchmarks studied, over 50\% of the binary variables have the same assignment across the positive sample and corresponding negative samples (Table \ref{tab:collect_stats}). By Proposition \ref{prop:1}, the CL loss provides zero gradient signal for these variables, which explains both why pure CL underperforms in PG/PI and why it has low feasibility rates. WCE, conversely, supervises every variable but lacks discriminative predictions where positive and negative samples differ. CL+WCE combines the two: CL where negatives differ from positives, and WCE on variables that share the same assignment in positive and negative samples. 

\begin{figure*}[htbp]%!htbp
   \centering
    \begin{subfigure}[htbp]{0.31\textwidth}
		\centering
		\includegraphics[width=0.99\textwidth]{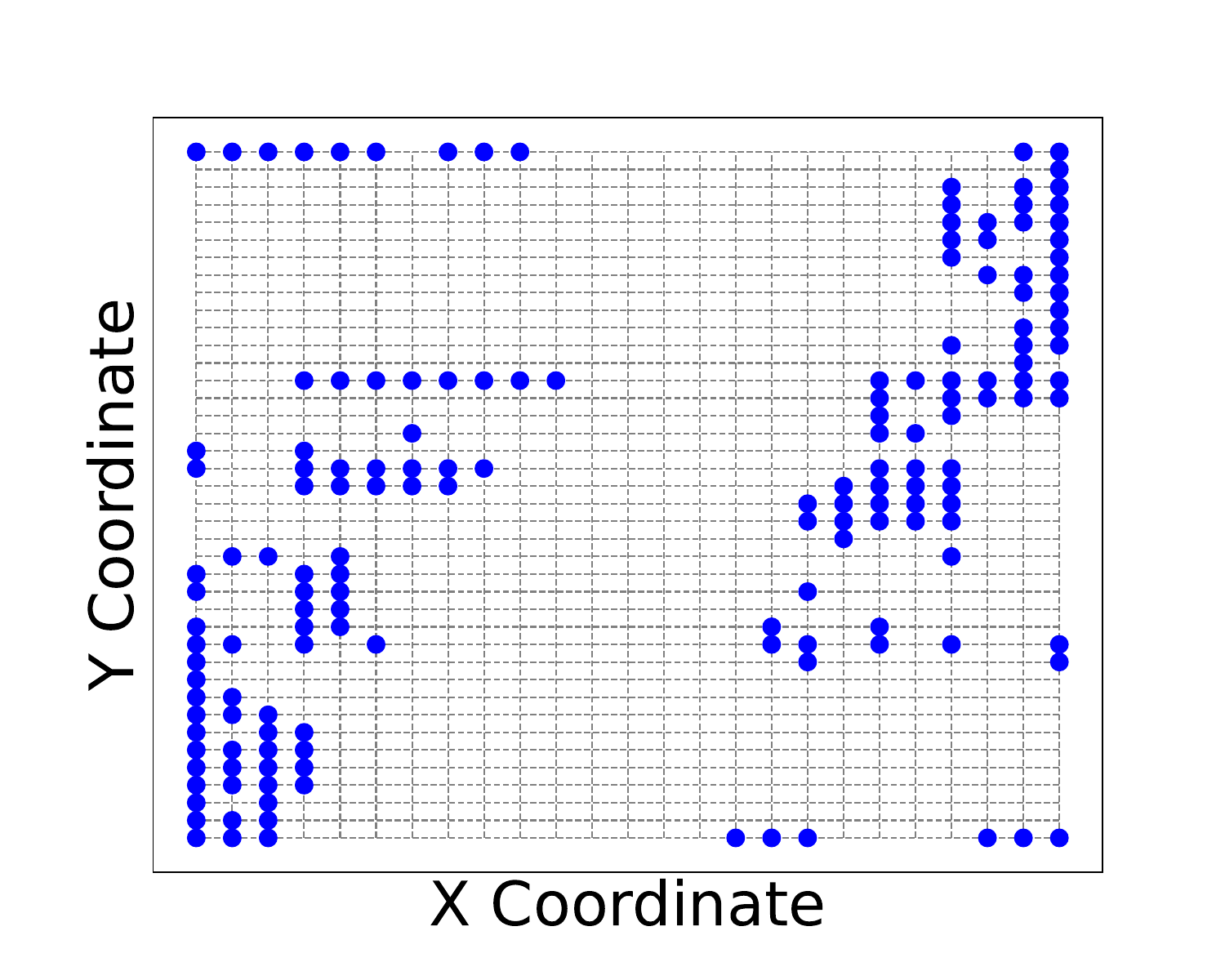}
		\caption{WCE}%\label{fig:1a}
		%\caption{An example with one agent.}
	\end{subfigure}
    \hspace{0em}
	\begin{subfigure}[htbp]{0.31\textwidth}
		\centering
		\includegraphics[width=0.99\textwidth]{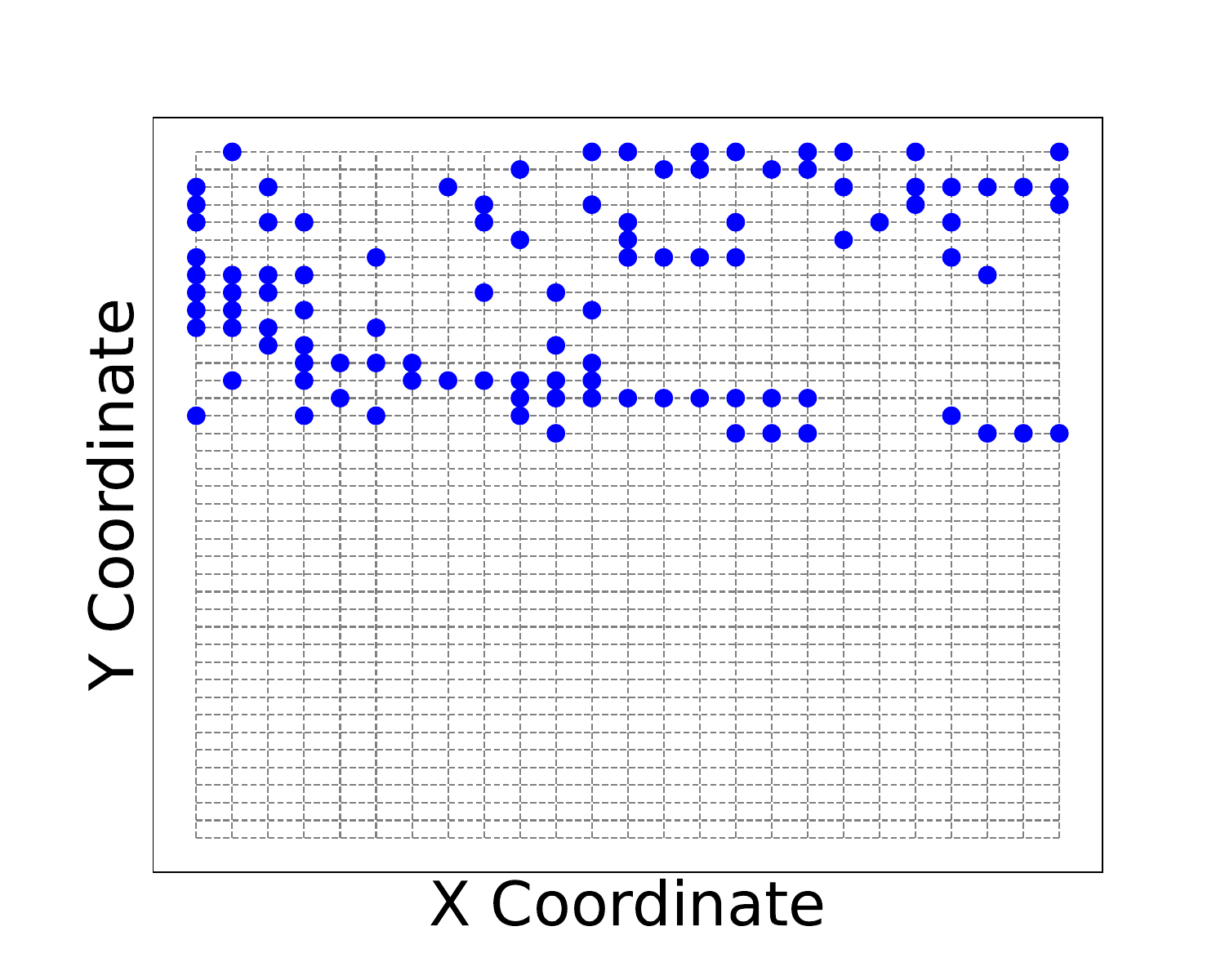}
			\caption{RENS}
	\end{subfigure}
    \hspace{0em}
	\begin{subfigure}[htbp]{0.31\textwidth}
		\centering
		\includegraphics[width=0.99\textwidth]{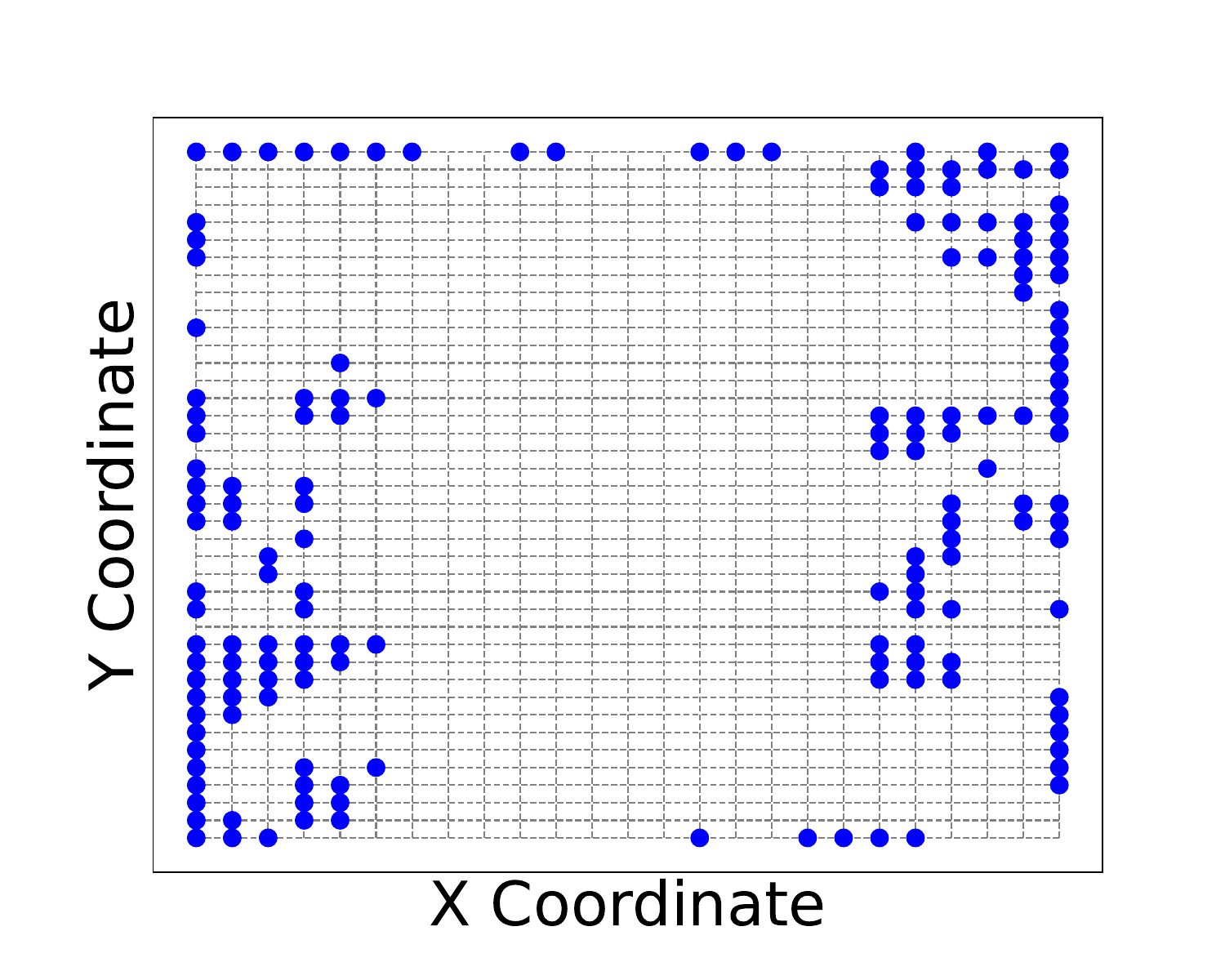}
			\caption{CL+WCE}
	\end{subfigure}

    \caption{Final wind farm layout solutions for a location in the Hawaii region predicted by different algorithms. Filled circles represent a wind turbine at that grid location.
    \label{fig:wflop}}
\end{figure*}

\begin{table*}[h]

\centering
\scalebox{0.85}{
\setlength{\tabcolsep}{5pt}
\begin{tabular}{llllllllllllll}
\toprule
                &            & \multicolumn{3}{c}{\bfseries CBQP} & \multicolumn{3}{c}{\bfseries QMKP} & \multicolumn{3}{c}{\bfseries CQKP} & \multicolumn{3}{c}{\bfseries WFLOP} \\
& \bfseries Method  & \bfseries PG & \bfseries PI &  \bfseries \# wins & \bfseries PG & \bfseries PI & \bfseries \# wins & \bfseries PG & \bfseries PI &  \bfseries \# wins & \bfseries PG  & \bfseries PI & \bfseries \# wins   \\
                              \midrule
Adapted &WCE & 0.01                   & 27.68                  & 4                              & \textbf{0.01}          & 23.34                  & 20                             & \textbf{0}             & 18.61                  & 34                             & 0.17                   & 44.57                  & 35                           \\
& CL   & 0.02                   & 24.05                  & 27                             & 0.68$^\dagger$                   & 45.68$^\dagger$                  & 16$^\dagger$                             & 0.33$^\dagger$                   & 27.82$^\dagger$                  & 21$^\dagger$                             & 0.42$^\dagger$                   & 50.84$^\dagger$                  & 17$^\dagger$  \\
 \midrule
Extended & CL+WCE     & 0.01                   & \textbf{23.48}         & \textbf{56}                    & \textbf{0.01}          & \textbf{20.79}         & \textbf{52}                    & \textbf{0}             & \textbf{18.43}         & \textbf{45}                             & \textbf{0.14}          & \textbf{43.84}         & \textbf{43}    \\

 \midrule
 \midrule
Baselines & Gurobi               & \textbf{0}             & 28                     & 13                             & 0.02                   & 23.86                  & 12                             & 0.02                   & 29.78                  & 0                              & 0.55                   & 54.83                  & 3    \\
& Relax-Search      & 0.65                   & 48.66                  & 0                              & 0.32                   & 39.46                  & 0                              & 0.76                   & 56.56                  & 0                              & 0.81                   & 59.18                  & 0   \\ 
& SurCo       & 1                      & 60                     & 0                              & 1                      & 60                     & 0                              & 0.99                   & 59.49                  & 0                              & 0.84                   & 55.08                  & 2   \\ 
\bottomrule 
\end{tabular}
}
\caption{
Ablation with Gurobi as the solver: Primal Gap (PG), Primal Integral (PI), and \# wins in terms of PI. Lower is better for PG and PI. Higher is better for \# wins. WCE and CL are ML-guided MBQP primal heuristics that we adapted from MILPs. CL+WCE is the extended ML method with the proposed combined loss function. $^\dagger$ indicates benchmarks where there are instances for which the method did not produce a feasible solution. For CL, the feasibility rates are 35\%, 78\%, and 97\% for QMKP, CQKP, and WFLOP. For all other methods and benchmarks, the feasibility rate is 100\%. Note that PG, PI, and \# wins apply to all instances regardless of feasibility. 
}\label{tab:res_grb}
\end{table*}

\section{Cross-regional Generalization on WFLOP}\label{cross}

The purpose of this study is to understand the impact of regional wind-pattern heterogeneity on the inference performance of the ML-guided primal MBQP heuristics. 
Wind farm layout optimization in offshore environments is greatly impacted by regional differences in wind resources caused by geographic factors such as proximity to shore, water depth, and local climate. 
This study attempts to understand the generalization capacity of ML-based primal heuristic methods to unseen regions.
The goal is to develop methods that generalize effectively, enabling the efficient design of wind farms using regional wind scenario distributions, even in regions with limited historical wind data for validation.
To quantify the cross-regional transferability, we fit the ML-guided primal heuristics using training instances across five U.S. regions and test on multiple unseen regions.

\paragraph{Setup} 
For the ML-guided solution methods, we train the ML models on a mix of 800 instances randomly selected from five regions: Pacific Northwest, Mid Atlantic, Maine, Gulf of Mexico, and South Atlantic. At inference time, we use the trained models to predict solutions for WFLOP instances from three unseen regions: California, Hawaii, and Great Lakes (100 test instances per region). 

\paragraph{Results and Discussion} 
As shown in Table \ref{tab:corssresults} and Fig. \ref{fig_primal_gap_tranfer}, RENS is a strong baseline for WFLOP. While the WCE-based ML method adapted from MILPs outperforms RENS in the in-domain settings of the previous section, it fails to do so in the cross-regional study. Our proposed extension, which combines the CL and WCE losses, improves cross-regional generalization and outperforms RENS---and all other baselines---in PG, PI, and \# wins in the California and Great Lakes regions. In the Hawaii region, CL+WCE does not surpass RENS in PI, but it outperforms RENS in PG and \# wins. One explanation for the ability of ML-guided methods to predict high-quality wind farm layouts in unseen regions is that the distributional summary of 10 years of simulated wind data can adequately capture the relevant attributes of offshore wind patterns across geographically distinct regions.
%This may be due to the spatiotemporal dependence between weather phenomena in the underlying wind simulation models used to generate the data. 

Additionally, we show an example of final wind farm layouts predicted by the best performing methods.  Fig.~\ref{fig:wflop} shows the solutions returned by the best performing methods in Table~\ref{tab:corssresults} in each category: adapted ML (WCE), extended ML (CL+WCE), and non-ML (RENS). 
The CL+WCE layout solution leads to a 17\% reduction in wind speed losses due to wake effects compared to the WCE layout solution (the second best solution).
Wind speed losses impact power production and are a function of turbine placement. 
And although it is also a function of the distribution of the wind scenarios, greater turbine spacing will often lead to lessened wake effects and less wind speed loss. 
Thus, the improvement in objective value in the design in Fig. \ref{fig:wflop}(c) may be attributed to the slightly more dispersed turbine clustering.

\section{Ablation with Gurobi}

To demonstrate that our methods are solver-agnostic, we conduct the same experiments using Gurobi \cite{gurobi}, a state-of-the-art commercial solver. For the adapted and extended ML-guided primal heuristics, Gurobi is used both to collect training data and to solve the sub-MBQPs at inference time. As baselines, we compare against standard Gurobi, as well as Relax-Search and SurCo implemented in Gurobi. We do not include RENS or Undercover, as Gurobi does not provide implementations of these methods. 
As shown in Table~\ref{tab:res_grb}, the extended CL+WCE method achieves the best PI and the highest number of PI wins in all four benchmarks. In terms of PG, CL+WCE performs best on three of the four benchmarks; the exception is CBQP, where Gurobi performs slightly better. This may be attributed to the strength of Gurobi as a commercial solver. Consistent with our observations in SCIP, the adapted CL-based method fails to produce feasible solutions for some instances in QMKP, CQKP, and WFLOP. These results confirm that the conclusions drawn from the SCIP experiments also carry over when switched to another solver.

\section{Conclusion and Discussion}
We develop ML-guided primal heuristics for MBQPs based on solution prediction. We adapt existing WCE- and CL-based methods, originally developed for ML-guided MILP primal heuristics, to the MBQP setting by introducing a tripartite graph representation, a neural network architecture for feature extraction, and a data collection procedure that produces diverse, high-quality solutions for training. In addition, we extend the existing loss functions used in CO solution prediction and combine the CL and WCE losses to address the lack of gradient signal in the CL-based method. Experimental results show that the adapted and extended ML-guided methods produce high-quality solutions quickly and converge faster than non-ML primal heuristics. Furthermore, our combined-loss extension significantly improves performance over other ML methods that use existing loss functions, and it improves generalization in cross-regional inference on WFLOP.

A potential limitation of our methods is their extensive memory requirements. The tripartite graph grows quadratically in size: it is $O(n+m+\rho n^2+mn)$, where $n$ is the number of variables, $m$ is the number of constraints, and $\rho$ is the quadratic term density (since each quadratic term node connects to two variable nodes). In addition, although the proposed ML method significantly outperforms non-ML primal heuristics at inference time, the offline data collection required to train it is computationally expensive.
%For future work, we plan to combine solution prediction and projection methods that restore feasibility.
%Mixed-Integer Quadratically Constrained Programs, a broader class where feasibility is more challenging

\section{Acknowledgments}
This work was done during Weimin Huang’s internship at the Pacific Northwest National Laboratory (PNNL) and at the University of Southern California. PNNL is a multi-program national laboratory operated by Battelle Memorial Institute for the U.S. Department of Energy (DOE) under Contract No. DE-AC05-76RL0-1830. The research is partially supported by the National Science Foundation (NSF) grant \#2112533: “NSF Artificial Intelligence Research Institute for Advances in Optimization (AI4OPT)”, the NSF grant \#2346058: “NRT-AI: Integrating Artificial
Intelligence and Operations Research Technologies", the U.S. Department of Energy, Office of Science Energy Earthshot Initiative, as part of the Addressing Challenges in Energy: Floating Wind in a Changing Climate Energy Earthshot Research Center at PNNL, and the Ralph S. O’Connor Sustainable Energy Institute (ROSEI) at Johns Hopkins University. 

%\bigskip

\bibliography{aaai2026}
\appendix

\clearpage
\section{MBQP Benchmarks}\label{inst}
\subsection{Formulation}
\subsubsection{Cardinality-constrained Binary Quadratic Programs (CBQP) \citep{zheng2012successive}}
\[
\min -\sum_{i=1,...,n} \sum_{j=1,...,n} q_{ij}x_i x_j \; 
\]
\[
\text{s.t.  } \sum_{i=1,...,n} x_i = k
\]
\[
x_i \in \{0, 1\}, \quad i = 1, \dots, n \\
\]
where $n$ is the number of binary variables and $q_{ij}$ are entries in a symmetric matrix. 
\subsubsection{Quadratic Multidimensional Knapsack Problem (QMKP) \citep{forrester2020computational}}

\[
\min -\sum_{i=1}^n \sum_{j=1}^n q_{ij} x_i x_j - \sum_{i=1}^n c_i x_i
\]
\[
\text{s.t.  } \sum_{i=1}^n a_{ik} x_i \leq b_k, \quad k = 1, \dots, m
\]
\[
x_i \in \{0, 1\}, \quad i = 1, \dots, n
\]

\subsubsection{Cardinality-constrained Quadratic Knapsack Problem (CQKP) \citep{letocart2014efficient}}
\[
\min -\sum_{i=1}^n \sum_{j=1}^n q_{ij} x_i x_j - \sum_{i=1}^n c_i x_i
\]

\[
\text{s.t.  } \sum_{i=1}^n a_{ik} x_i \leq b_k, \quad k = 1, \dots, m
\]
\[
\sum_{i=1,...,n} x_i = K
\]
\[
x_i \in \{0, 1\}, \quad i = 1, \dots, n \\
\]

\subsubsection{Wind Farm Layout Optimization Problem (WFLOP) \citep{Huangsocs_2025}}

Let $J$ be the set of candidate locations for turbine placement and $K$ be the number of turbines to be installed. The binary decision variable $y_j$ takes the value 1 if a wind turbine is installed at location $j$, and 0 otherwise. The set $J$ is a two-dimensional grid with a given resolution that represents the design area of the wind farm. Let $U$ and $\theta$ represent the free stream wind speed and wind direction, respectively. Let $M$ be a set of wind scenarios $\mathcal{M} = \{1,2,\ldots,M\}$ drawn from a joint probability distribution $p(U,\theta)$. Each scenario $m$ consists of a wind speed $U^{(m)}$ and wind direction $\theta^{(m)}$ with  probability $p^{(m)}$ such that $\sum_{m \in \mathcal{M}} p^{(m)}  = 1$. 
The pairwise wind speed deficit interactions $d_{ij}^{(m)}$ are a function of additional parameters, including wind direction $\theta^{(m)}$. WFLOP MBQP minimizes expected wind speed losses. 

\[
\text{min} \quad  \sum_{m \in \mathcal{M}} p^{(m)}\, U^{(m)} \sum_{j \in J}\sum_{i \in J} \left(d_{ij}^{(m)}\right)^2\, y_i  y_j 
\]
\[
\text{s.t.} \quad  \sum_{j \in J} y_j = K 
\]
\[
y_j \in \{0,1\} \quad \forall j \in J
\]

\subsection{Instance statistics}
The instance statistics are shown in Table \ref{tab-stas}.
\begin{table} [!htbp]
\centering

%\begin{tabularx}{0.8\textwidth} { 
%  | >{\centering\arraybackslash}X 
%  | >{\centering\arraybackslash}X 
%  | >{\centering\arraybackslash}X 
%  | >{\centering\arraybackslash}X | }
%\scriptsize
\scalebox{0.9}{
\begin{tabular}{|r|r|r|r|r|}
 \hline
Category & Benchmark & \# Var & \# Cons & Quad. den. \\
 \hline
Synthetic &  CBQP & 1000 &  1& 25.00\% \\
 \hline
Synthetic & QMKP  & 1000 &  5 &   25.00\% \\
 \hline
Synthetic & CQKP & 1000 &  2 &  25.00\% \\
 \hline
Real-world & WFLOP & 1000 &  1 &  32.42\% \\
 \hline

\end{tabular}}
\caption{Instance statistics. Number of binary variables (\# Var), number of constraints (\# Cons), and quadratic term density (Quad. den).}
\label{tab-stas}
\end{table}

\section{Proof of Proposition}\label{pr_prop}
\begin{figure*}[t]
\begin{equation}
	\begin{aligned}
    \exp \left({x^+}\cdot p(\theta;{M_i}) / \tau(x^+|M_i)\right)
=\exp \left(({x}^{+^{U^{x^+}_i}}\cdot p(\theta;{M_i})^{\mathcal{U}_{i}^{x^{+}}} + {{x}^{+}}^{\mathcal{B}_i\setminus {\mathcal{U}_{i}^{x^{+}}}} \cdot p(\theta;{M_i})^{\mathcal{B}_i\setminus {\mathcal{U}_{i}^{x^{+}}}}) / \tau(x^+|M_i)\right) \\
 = \exp \left({x}^{+^{\mathcal{U}_{i}^{x^{+}}}}\cdot p(\theta;{M_i})^{\mathcal{U}_{i}^{x^{+}}} / \tau(x^+|M_i)\right) \exp \left({x}^{+^{\mathcal{B}_i\setminus {\mathcal{U}_{i}^{x^{+}}}}}\cdot p(\theta;{M_i})^{\mathcal{B}_i\setminus {\mathcal{U}_{i}^{x^{+}}}} / \tau(x^+|M_i)\right)
	\end{aligned}\label{proof-eqn1}
\end{equation}

%%%%%%%
\begin{equation}
	\begin{aligned}
\exp \left({\Tilde{x}}\cdot p(\theta;{M_i}) \right/ \tau(x^+|M_i))=\exp \left((\Tilde{x}{^{U^{x^+}_i}}\cdot p(\theta;{M_i})^{\mathcal{U}_{i}^{x^{+}}} + {\Tilde{x}}^{\mathcal{B}_i\setminus {\mathcal{U}_{i}^{x^{+}}}} \cdot p(\theta;{M_i})^{\mathcal{B}_i\setminus {\mathcal{U}_{i}^{x^{+}}}}) / \tau(x^+|M_i)\right) \\
 = \exp \left({\Tilde{x}}^{{\mathcal{U}_{i}^{x^{+}}}}\cdot p(\theta;{M_i})^{\mathcal{U}_{i}^{x^{+}}} / \tau(x^+|M_i)\right) \exp \left({\Tilde{x}}{^{\mathcal{B}_i\setminus {\mathcal{U}_{i}^{x^{+}}}}}\cdot p(\theta;{M_i})^{\mathcal{B}_i\setminus {\mathcal{U}_{i}^{x^{+}}}} / \tau(x^+|M_i)\right)
	\end{aligned}\label{proof-eqn2}
\end{equation}

\begin{equation}
	\begin{aligned}
	\mathcal{L}^+\left({\theta} \mid x^+, {M_i}\right)
        &=- \log \frac{ \exp \left({x^{+}}^{\mathcal{B}_i\setminus {\mathcal{U}_{i}^{x^{+}}}}\cdot p(\theta;{M_i})^{\mathcal{B}_i\setminus {\mathcal{U}_{i}^{x^{+}}}} / \tau(x^+|M_i)\right)}{\sum_{\Tilde{x} \in S^-_{M_i} \cup \{x^+\}} \exp \left(\Tilde{x}^{\mathcal{B}_i\setminus {\mathcal{U}_{i}^{x^{+}}}}\cdot p(\theta;{M_i})^{\mathcal{B}_i\setminus {\mathcal{U}_{i}^{x^{+}}}} / \tau(x^+|M_i)\right)}.
	\end{aligned}\label{proof-eqn3}
\end{equation}
%%%%%%

\end{figure*}

Let $\theta$ be the set of weights learned by an ML model and $p(\theta;{M_i})$ be the vector of variable predictions produced by the model. We use $p(\theta;{M_i})^{\mathcal{U}_{i}^{x^{+}}}$ to denote the prediction vector for the variable subset ${\mathcal{U}_{i}^{x_{+}}}$, (i.e., $[p_{j}(\theta;{M_i})]_{\forall j\in {\mathcal{U}_{i}^{x^{+}}}}$). Similarly, we use $p(\theta;{M_i})^{\mathcal{B}_i\setminus {\mathcal{U}_{i}^{x_{+}}}}$ to denote the prediction vector for the subset ${\mathcal{B}_i\setminus {\mathcal{U}_{i}^{x_{+}}}}$. We use $\tilde{x}^{\mathcal{U}_{i}^{x^{+}}}$ to denote the variable assignments in ${\mathcal{U}_{i}^{x^{+}}}$ in a solution $\tilde{x}$ (i.e., $[\tilde{x}_j]_{\forall j\in {\mathcal{U}_{i}^{x^{+}}}}$).

%$p(x|\theta;{M_i})$ is a vector of predictions,
%$\omega_i = [p{,j}(x|\theta;{M_i})]_{\forall j\in {\mathcal{U}_{i}^{x^{+}}}}$. 
%Let $x^{\mathcal{U}_{i}^{x^{+}}}$ and $p(\theta;{M_i})^{\mathcal{U}_{i}^{x^{+}}}$ denote vectors of solution $x$ and predictions $p(\theta;{M_i})$ with variables in ${\mathcal{U}_{i}^{x^{+}}}$. Let $x^{\mathcal{B}_i\setminus {\mathcal{U}_{i}^{x^{+}}}}$ and $p(\theta;{M_i})^{\mathcal{B}_i\setminus {\mathcal{U}_{i}^{x^{+}}}}$ denote a vector set that contains the unchanged variables. 
By the property of the dot product operation, $\exp \left({x^+}\cdot p(\theta;{M_i}) / \tau(x^+|M_i)\right)$ can be decomposed as shown in Equation 
(\ref{proof-eqn1}). Similarly, $\exp \left({\Tilde{x}}\cdot p(\theta;{M_i}) \right/ \tau(x^+|M_i))$ can be decomposed as shown in Equation 
(\ref{proof-eqn2}). By definition of the set $U^{x^+}_i$, ${\Tilde{x}}^{{\mathcal{U}_{i}^{x^{+}}}}={x^+}^{{\mathcal{U}_{i}^{x^{+}}}}$. Combining Equations (\ref{proof-eqn1}), (\ref{proof-eqn2}), and (\ref{eq:cl}) leads to Equation (\ref{proof-eqn3}). Therefore, $\mathcal{L}^+\left({\theta} \mid x^+, {M_i}\right)$ is a function of $p(\theta;{M_i})^{\mathcal{B}_i\setminus {\mathcal{U}_{i}^{x^{+}}}}$.

%However, there has been limited existing work on primal heuristics specifically designed for solving MBQPs - many primal heuristics were derived from algorithms for Mixed Integer Linear Programming (MILP) problems \citep{berthold2014rens,bonami2009feasibility}.

\section{Numerical Stability}
To avoid numerical overflow issues, we follow existing work on WCE-based MILP solution prediction in \cite{han2023a} which divides the objective value by a constant $\lambda$ (which is known as the weight norm parameter) before computing the exponentials. In practice, the weight applied to sample $x^+$ for instance $M_i$ is
$$
			 w(x^+|M_i) \equiv \frac{\text{exp}\left((-{x^{+\top} H_i x^+}-{c_i}^{\top} x^+)/\lambda\right)}{\sum_{{x} \in \mathcal{S}^{+}_{M_i}}\exp{ \left((-x^{\top} H_i x-{c_i}^{\top}x)/\lambda\right)}}. $$
             
%Implementation of \cite{han2023a} is publicly availabe at \url{https://github.com/sribdcn/Predict-and-Search_MILP_method/}. 
$\lambda$ is a positive constant in instances from the same benchmark. Since \cite{han2023a}'s implementation only provides hard-coded $\lambda$ values for each benchmark but did not provide details on how the values were selected, we select a $\lambda$ value so that $|-\text{avg. obj}/\lambda| = 10$ to avoid extremely large terms in computing the exponentials, where avg. obj is the average objective value across $S^+$ (the set of good solutions) and across instances in the training data for each benchmark. 

\section{Neural Network Architecture}
\paragraph{List of features}
The full list of features for the tripartite graph is shown in Table \ref{features}. 

\begin{table*}[h]
\centering
\scalebox{0.9}{
\begin{tabular}{llll}
\toprule
Nodes & Features & Source \\
\hline\hline
C  & avg. coefficients in the constraint                                  & \citep{han2023a}  \\
& min. coefficients in the constraint                                  & new  \\
& max. coefficients in the constraint                                  & new  \\
    & variance of coefficients in the constraint                                  & new  \\
                    &  \# of variables in the constraint   & \citep{han2023a}  \\
                    & left-hand side or right-hand side                                         &  \citep{han2023a}  \\
                    & constraint sense in one-hot encoding (3) $(=,>,<)$                                  & new  \\
                    \midrule
V-C edge        & coefficient of variables in constraints    & \citep{han2023a}  \\
\midrule
V    & normalized coefficient in obj (among linear terms)              & \citep{han2023a}  \\
                    & avg. coefficient in constraints                     & \citep{han2023a}  \\
                    & \# of times it appears in linear constraints                        & \citep{han2023a}  \\
                    & variance of coefficient in constraints                       & new  \\
                    & max. coefficient in constraints                         & \citep{han2023a}  \\
                    & min. coefficient in constraints                         & \citep{han2023a}  \\
                    & binary variable indicator          & \citep{han2023a}  \\
                    & LP relaxation value in MILP reformulation  & new \\
                    & \# times it appears in quadratic terms  &   new \\
                    & avg. coefficient in quadratic terms that it appears in &  new \\
                    & max. coefficient in quadratic terms that it appears in &  new \\
                    & min. coefficient in quadratic terms that it appears in &  new \\
                    & variance of coefficient in quadratic terms that it appears in &   new \\     
                    & avg. \# times its neighbors appear in quadratic terms &  new \\
                    & max. \# times its neighbors appear in quadratic terms &  new \\
                    & min. \# times its neighbors appear in quadratic terms &  new \\
                    & variance of \# times its neighbors appear in quadratic terms &   new \\    
                    & eigenvector centrality in Hessian graph &  new \\
\midrule
Q    & coefficient of quadratic term in objective function      & new  \\
& LP relaxation value of reformulated variable $z_{ij} = x_i x_j$      &  new  \\
& LP relaxation violation      &  new  \\
& Edge centrality in Hessian graph     &  new  \\
                    \midrule
V-Q edge        & None         &  new  \\
\bottomrule
\end{tabular}}
\caption{Features of MBQP tripartite graph representation.}
\label{features}
\end{table*}
\end{document}